\documentclass[iicol,referee]{sn-jnl}

\usepackage{graphicx}%
\usepackage{multirow}%
\usepackage{amsmath,amssymb,amsfonts}%
\usepackage{amsthm}%
\usepackage{mathrsfs}%
\usepackage[title]{appendix}%
\usepackage{xcolor}%
\usepackage{textcomp}%
\usepackage{manyfoot}%
\usepackage{booktabs}%
\usepackage{algorithm}%
\usepackage{algorithmicx}%
\usepackage{algpseudocode}%
\usepackage{listings}%
\usepackage{comment}
\usepackage[utf8]{inputenc} 
\usepackage{graphicx}       
\usepackage{amsmath}  
\usepackage{multirow}
\usepackage{booktabs}
\usepackage{array} 

\theoremstyle{thmstyleone}%
\theoremstyle{thmstyletwo}%

\theoremstyle{thmstylethree}%

\newcommand{\Florent}{\textcolor{black}}
\newcommand{\Simon}{\textcolor{black}}

\begin{document}

\title{Domain Adaptation Based Pipeline for Character Classification and Handwritten \Florent{Text} Recognition}








\author{\fnm{Florent} \sur{Imbert}}\email{florent.imbert@associated.ltu.se}
\equalcont{These authors contributed equally to this work.}

\author{\fnm{Simon} \sur{Corbillé}}\email{simon.corbille@associated.ltu.se}
\equalcont{These authors contributed equally to this work.}

\author{\fnm{Hui} \sur{Han}}\email{hui.han@ltu.se}

\author{\fnm{Elisa H.} \sur{Barney Smith}}\email{elisa.barney@ltu.se}

\affil{\orgdiv{Machine Learning Group}, \orgname{Luleå University of Technology}, \orgaddress{\city{Luleå}, \country{Sweden}}}


\abstract{
Character classification and handwritten text recognition, particularly in specialized tasks, often suffer from limited labeled data, while a large amount of unlabeled data remains unused. To address this challenge, we propose a novel unsupervised domain adaptation-based pipeline to reduce the distribution gap between source and target domains. 
Our pipeline follows three main steps. In the first step, we pretrained an expert model on source data, to learn a strong general representation. In the second step, we perform unsupervised domain adaptation to leverage unlabeled target data, adapting the model to the specific characteristics of the target domain. 
In the third step, we trained the expert model on source data in the target style.
By aligning the feature distributions between the source and target, our method enables the model to generalize effectively across diverse datasets.
The unsupervised training increases the computation time only during training not in inference.
We evaluate the pipeline on both character-level and historical handwritten line recognition tasks, demonstrating its flexibility.
We establish a new state-of-the-art on the Safran-MNIST-DLS dataset defined in Track~2 of the DAGECC competition with 76.17\% Macro F1-score and achieve competitive results on each of the specific datasets of READ 2018 without using the labels of the target data. 
Code is available at \url{https://github.com/simon-corbi/DRANet-for-classification-and-htr}}




\maketitle

\section{Introduction}\label{sec_intro}
Large-scale image datasets from both public online sources and industrial environments exist and provide diverse visual content such as scanned documents, natural scene images, and born-digital materials. In this work, we focus on images of text, where typically only a small portion of the data is annotated, while the majority remains unlabeled. The annotations themselves can take different forms, for example, objects present in an image or the transcription of the text it contains. 
Depending on the task and the available resources, deep neural networks can be trained in a fully supervised manner using labeled data or in an unsupervised manner using unlabeled data. However, a well-known limitation of supervised learning is that models often generalize poorly when applied to unseen data that follow a different distribution from the training data \cite{domain_shift}.

Data annotation requires significant time, financial cost, and expert knowledge. This makes it impractical to build large annotated datasets in many real-world scenarios. A promising alternative is to exploit unlabeled data from the target domain to adapt a model that was initially trained on labeled source domain data. This approach, known as domain adaptation, is a branch of transfer learning that aims to reduce the distributional gap between source and target data, thereby improving performance on the target domain. By leveraging large amounts of unlabeled data, domain adaptation provides a way to build more robust models without additional manual labeling effort. 
We use domain adaptation for two tasks:  character classification and handwritten text recognition (HTR). 
In addition to the academic interest, these tasks have practical value in areas such as serial number recognition for industrial quality control and automated document analysis for digital archiving.

\begin{figure}[t]
\centering  
    \includegraphics[width=\linewidth]{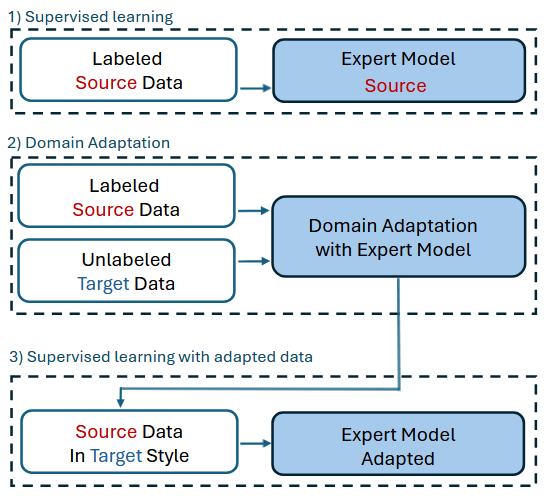}
\caption{Overview of the proposed pipeline. Step 1): supervised training of the expert model on source-domain data. Step 2): unsupervised domain adaptation using DRANet to generate adapted data in the target style. Step 3): training of the expert model on the adapted data.}
\label{fig_overview_method}
\end{figure}

In this work, we propose a novel domain adaptation pipeline for character classification and handwritten recognition illustrated in Figure~\ref{fig_overview_method}. We begin by designing and training an expert model on labeled source data. Secondly, we jointly train this expert model with a domain adaptation module using both labeled source data and unlabeled target data. Thirdly, we trained the expert model with source data in target style in order to aim to align feature representations and transfer knowledge across domains.

To demonstrate the generality of our approach, we evaluate it on two tasks: character classification and handwritten text recognition. 

We evaluate the effectiveness of our proposed pipeline against a fully supervised model and show that it achieves competitive results without requiring any additional annotations.

The main contributions of this work are as follows:
\begin{itemize}
    \item Introduction of a novel pipeline to perform character classification and handwritten text recognition, which includes an expert model and a domain adaptation module based on DRANet.
    \item A synthesis method to generate character images resembling serial numbers engraved on aircraft parts.
    \item An ablation study to compare the impact of training data (labeled, unlabeled) in supervised learning and domain adaptation learning settings. 
    \item Comparison of our method against the state-of-the-art in two tasks. 
\end{itemize}

This paper is organized as follows. Section~\ref{sec_soa} describes the related works. Section~\ref{sec_method} details our method. Section~\ref{sec_expe} presents the quantitative and qualitative results. Then, we conclude in section~\ref{sec_conclu} and outline future research directions.

\section{Related Work}\label{sec_soa}
\subsection{Character Classification}
Character classification can be defined as the task of recognizing the isolated character contained in an image.
The character can be a digit~\cite{MNIST}, a letter~\cite{EMNIST} or a punctuation symbol.
It can be printed, handwritten~\cite{CARDIS,Mozaffari09, Seuret23} or engraved~\cite{Prakash23}.
In this work, to evaluate our Domain Adaptation technique for character classification, we focus on data from Track~2 of Domain Adaptation and Generalization for Character Classification (DAGECC)~\cite{DAGECC}, an ICPR 2024 competition which tackled unsupervised domain adaptation.
The Safran-MNIST-DLS dataset associated with Track 2 contains 32~classes depicting digits, alphabetic characters, and symbols (dot and slash) representing serial numbers on manufactured objects displayed in an engraved dot style.
The challenge is to recognize characters in a dataset where no labels are available to train a network. 
Most methods submitted to the competition are based on a ResNet~\cite{ResNet} Convolutional Neural Network~(CNN) pretrained on ImageNet~\cite{ImageNet} and used the EMNIST character dataset~\cite{EMNIST} together with custom synthetic data for the training.
It is worth noting that the top three methods did not utilize the unlabeled data from Safran-MNIST-DLS. In contrast, our approach explicitly leverages this unlabeled data through a domain adaptation framework to enhance ResNet’s performance.

\subsection{Handwritten Text Recognition}
The current state-of-the-art for handwritten text recognition at the line level is achieved by Convolutional Recurrent Neural Networks (CRNNs)~\cite{Retsinas22,Tarride24,Corbille25} and Transformers~\cite{barrere24,LiL23}.
These networks are trained in a supervised manner.
Since Transformers are generally slower and require larger amounts of training data than CRNNs, we use a CRNN in this work to address the handwritten text recognition task.

\subsection{Domain Adaptation}\label{sec_sota_DA}
Domain adaptation (DA) addresses the challenge of transferring knowledge from a source domain, where labeled data are abundant, to a target domain, where labels are scarce or unavailable. The central goal is to mitigate domain shift, \textit{i.e.} the discrepancy in data distributions between source and target, so that models trained on the source domain can generalize effectively to the target. 

Depending on the availability of target domain labels, three scenarios can be distinguished. In the supervised case, labeled target data are available. In the semi-supervised case, both labeled and unlabeled target data are provided. In the unsupervised case, only unlabeled target data are accessible.

DA techniques are typically grouped into three main categories: divergence-based, adversarial-based, and reconstruction-based approaches.

Divergence-based DA aims to minimize statistical discrepancies between domains by aligning their feature distributions. Metrics such as Maximum Mean Discrepancy and Kullback–Leibler divergence are often employed to reduce the inter-domain gap and encourage transferable representations \cite{shen2018wassersteindistanceguidedrepresentation}.

Adversarial-based DA adopts the adversarial training paradigm from Generative Adversarial Networks, where a feature extractor and domain discriminator engage in a minimax game to generate domain-agnostic feature representations \cite{DANN,DA_IMU}. Through this adversarial process, the feature extractor learns to fool the domain discriminator by producing representations that are invariant to domain-specific characteristics, effectively bridging the distributional gap between domains.

Reconstruction-based DA leverages encoder-decoder frameworks to learn domain-invariant features by imposing reconstruction fidelity as a regularization constraint \cite{ghifary2016deepreconstructionclassificationnetworksunsupervised}. The underlying principle dictates that meaningful feature representations should enable faithful reconstruction of input data irrespective of domain origin, thereby fostering cross-domain transferability and model generalization.

These complementary strategies synergistically advance DA by addressing different facets of the domain shift problem, distributional alignment, feature invariance, and representational robustness, resulting in models that demonstrate superior generalization capabilities and consistent performance across varied operational contexts.

In the context of character classification, DA addresses the significant challenge of transferring knowledge across diverse textual environments with dramatically varying visual properties. Character classifiers encounter substantial domain shifts when transitioning between different document types, writing styles, image acquisition conditions, and textual formats. These variations manifest as differences in font families, handwriting styles, document degradation levels, background textures, lighting conditions, noise, and image resolution, all of which can severely impact classification accuracy when models trained on one domain are applied to another. The complexity of DA for character classification is further amplified by the hierarchical nature of text recognition, which requires robust feature extraction at multiple levels, from low-level stroke patterns and character shapes to high-level contextual understanding. Consequently, effective DA strategies for character classification must simultaneously address visual feature alignment and preserve semantic consistency across domains to maintain both character-level accuracy and contextual coherence.

One common DA method is Deep Joint Distribution Optimal Transport (DeepJDOT) \cite{DeepJDOT} which employs optimal transport theory to align joint distributions of features and labels between source and target domains. This method ensures that the learned representations are not only domain-invariant but also preserve the discriminative information necessary for accurate classification. By minimizing the discrepancy between the joint distributions, DeepJDOT effectively bridges the domain gap, leading to improved character classification performance on target domain data.

\begin{figure*}[ht]
\centering  
    \includegraphics[width=\linewidth]{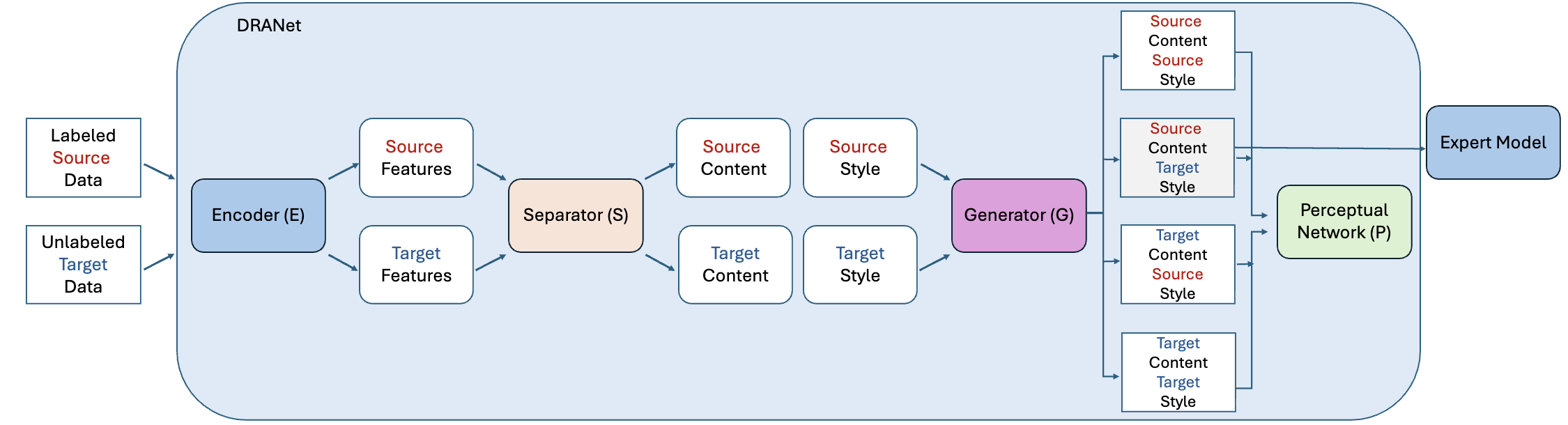}
\caption{Overview of the DRANet architecture. \Florent{Colored background round boxes represent networks, rectangular boxes represent images, white background round boxes represent features (features, content, style)}. The network is composed of an encoder (E), a separator (S), and a generator (G). Given labeled source images and unlabeled target images, the encoder E extracts the corresponding source features and target features. The separator S disentangles each feature into two components: the content and the style. The generator G then recombines the content and style representations to reconstruct or translate images, which are then used to calculate adversarial, reconstruction, style/content consistency, and perceptual losses through the perceptual network (P). \Florent{Note: only the Expert Model is used during inference.}}
\label{fig_DRANET}
\end{figure*}

Variations in visual appearance can significantly hinder a model’s ability to learn robust representations. Disentangling Representation and Adaptation Networks (DRANet) \cite{DRANet} introduces a novel architecture that disentangles content and style representations in images. In character classification, this separation isolates the structural information (content) of text from its visual appearance (style) (\textit{cf.} Figure~\ref{fig_DRANET}).

DRANet adapts the domain by transferring the style from the target domain to the content of the source domain, thereby preserving the inherent structure of the text while accommodating domain-specific visual variations. This approach has shown state-of-the-art performance in unsupervised cross-domain adaptation, making it highly effective for character classification applications across diverse datasets.


\Florent{In summary, DRANet \cite{DRANet} outperforms DeepJDOT \cite{DeepJDOT} for character classification thanks to its ability to disentangle content and style, preserving text structure while adapting to diverse visual appearances. Its flexible design allows us to extend DRANet to line-level Handwritten Text Recognition by introducing task-specific modifications to the encoder and classifier, addressing sequential dependencies and recognition requirements. These adjustments leverage the original adaptation pipeline, enabling robust cross-domain performance even in the presence of pronounced style variations.}

\section{Method}\label{sec_method} 
We propose a new pipeline to \textit{jointly} train an expert model for a main task in a supervised manner and a DA model for a DA task in an unsupervised manner. 
The overall structure of the proposed pipeline is illustrated in Figure~\ref{fig_overview_method}.
The expert model serves two purposes: it performs the main recognition task and provides feature representations for the DA model.
In our work, we use this pipeline for the main tasks of character classification and handwritten recognition.
We first present the expert models used and then the DA model.

\subsection{Expert Models}
\subsubsection{Convolutional Neural Network}\label{sub_sec_cnn} 
For the character classification task, we use a ResNet-34 \cite{ResNet} Convolutional Neural Network as the expert model.
The principal advantage of this architecture is that it includes residual connections to enhance the training and improve the accuracy of the model.
The architecture is detailed in Table~\ref{tab_detail_resnet}.



\begin{table}[t]
\caption{Architecture of ResNet-34 for image of size 48 x 48, with the numbers of blocks stacked. It is our expert model for the character classification task.}
\centering
\begin{tabular*}{\linewidth}{ccc}  
\cmidrule{2-2}
~~~~~~~~~~~~~~~~~~ & \textbf{Configuration} &   \\ \cmidrule{2-2}
& 7×7, 64, stride 2 &     \\ \cmidrule{2-2}
& 3×3 Max Pool, stride 2   \\ \cmidrule{2-2}
& $\left[\begin{array}{c} 3 \times 3,\, 64 \\ 3 \times 3,\, 64 \end{array}\right] \times 2$ & \\ \cmidrule{2-2}
& $\left[\begin{array}{c} 3 \times 3,\, 128 \\ 3 \times 3,\, 128 \end{array}\right] \times 4$ & \\ \cmidrule{2-2}
& $\left[\begin{array}{c} 3 \times 3,\, 256 \\ 3 \times 3,\, 256 \end{array}\right] \times 6$ & \\ \cmidrule{2-2}
&$\left[\begin{array}{c} 3 \times 3,\, 512 \\ 3 \times 3,\, 512 \end{array}\right] \times 3$ & \\ \cmidrule{2-2}
&average pool   &            \\ 
& 32-d fc, softmax  &               \\ 
\cmidrule{2-2}
\end{tabular*}
\label{tab_detail_resnet}
\end{table}

In addition to the cross-entropy loss, the model incorporates a center loss ~\cite{Wen016} to improve feature discrimination. The centers are computed every epoch on the whole training set.
The total loss is the sum of the cross-entropy and the center loss, where both losses have the same weight.

\subsubsection{Convolutional Recurrent Neural Network}\label{sub_sec_crnn}
For the handwritten text recognition task, we use the architecture defined in~\cite{Retsinas22} and extended in~\cite{Corbille25} where added a center loss to improve the performance.
The model illustrated in Table \ref{tab_detail_crnn} is composed of a Convolutional Neural Network (CNN) backbone followed by a recurrent head composed of BLSTM layers.
We add a squeeze and excitation block~\cite{SELayer} after each convolution block.
The network is trained with the Connectionist Temporal Classification (CTC) method~\cite{Graves06} and the center loss defined in~\cite{Corbille25}.

\begin{table}[t]
\caption{Architecture of CRNN. It is our expert model for the handwritten text recognition task.}
\centering
\begin{tabular*}{\linewidth}{>{\centering\arraybackslash}p{0.70\linewidth}|>{\centering\arraybackslash}p{0.15\linewidth}}
\toprule
\textbf{Main} & \textbf{Shortcut} \\ \midrule
7$\times$7, 64, stride 2 & \\ \cmidrule(lr){1-1}
3$\times$3 Max Pool, stride 2 & \\ \cmidrule(lr){1-1}
$\left[\begin{array}{c}3\times3,\,64\\3\times3,\,64\end{array}\right]\times2$ & \\ \cmidrule(lr){1-1}
Squeeze and excitation & \\ \cmidrule(lr){1-1}
2$\times$2 Max Pool & \\ \cmidrule(lr){1-1}
$\left[\begin{array}{c}3\times3,\,128\\3\times3,\,128\end{array}\right]\times3$ & \\ \cmidrule(lr){1-1}
Squeeze and excitation & \\ \cmidrule(lr){1-1}
2$\times$2 Max Pool & \\ \cmidrule(lr){1-1}
$\left[\begin{array}{c}3\times3,\,256\\3\times3,\,256\end{array}\right]\times4$ & \\ \cmidrule(lr){1-1}
Squeeze and excitation & \\ \cmidrule(lr){1-1}
Column Max Pool & \rotatebox[origin=c]{270}{$\Rsh$} \\ \midrule
BiLSTM, 256$\times$3 & $3\times$ n classes \\ \midrule
Linear, n classes & \\ \cmidrule(lr){1-1}
\end{tabular*}
\label{tab_detail_crnn}
\end{table}

\subsection{Domain Adaptation}
\label{sub_DA}

We adopt the Disentangling Representation and Adaptation Network (DRANet) \cite{DRANet}, which explicitly separates content (textual structure) from style (visual appearance). This property makes DRANet particularly suitable for character classification and handwritten text recognition tasks, as it ensures structural consistency of the recognized text while adapting to diverse styles.
\Florent{DRANet is designed for unsupervised domain adaptation and relies exclusively on labeled source-domain data and unlabeled target-domain data during training. The target samples are used to guide the domain alignment process and to generate adapted images that preserve semantic content while capturing target-domain appearance, without requiring any target annotations.}
\Florent{However, applying the generic DRANet architecture without task-specific adjustments is not sufficient for the considered recognition task. 
Therefore, we introduce task-specific adjustments to the DRANet pipeline, illustrated in Figure~\ref{fig_DRANET}. Our main contributions lie in the design of a task adapted Encoder (E) and a dedicated Expert Model that jointly improve domain-invariant feature learning and recognition performance. For each task, we refine state-of-the-art models architectures, using the initial layers as the Encoder (E) and the full refined model as the Expert Model. This approach balances strong feature extraction capabilities with task-specific recognition performance.
The pipeline is as follows:}
\begin{itemize}
    \item \textbf{Encoder}(E): The encoder receives an input image (from either the source or target domain) and maps it to a latent representation. In this work, the encoder is made task-specific by utilizing either the convolutional component of ResNet-34 \textit{cf.} Section \ref{sub_sec_cnn} or the feature extraction layers preceding the linear classification stage in the CRNN \textit{cf.} Section \ref{sub_sec_crnn}.
    \item \textbf{Separator}(S): The separator divides the latent representation into content and style. It is implemented as a neural module comprising two consecutive 2D convolutional layers, each with a kernel size of 3×3, stride 1, and padding 1, followed by ReLU activations. Spectral normalization is applied to both convolutional layers to stabilize training. During the forward pass, style features are obtained by passing the latent representation through the convolutional layers, while content features are computed by subtracting the style features from the original latent representation.
    \item \textbf{Generator}(G): The generator reconstructs images from the combined content and style representations and can also produce cross-domain images by merging content from one domain with style from another. The network is implemented as a sequence of upsampling and convolutional operations. Specifically, it consists of three bilinear upsampling layers, each doubling the spatial resolution of the input, followed by spectrally normalized transposed convolutional layers with kernel size 3×3 and stride 1, and ReLU activations. The final layer is a 3×3 convolution followed by a Tanh activation to produce the reconstructed image. During the forward pass, content and style representations are summed and processed through the upsampling–convolution sequence. The output is further interpolated to a fixed spatial size, ensuring consistent image dimensions across the dataset.
    \item \textbf{Perceptual Network (P)}: The perceptual network ensures that reconstructed or adapted images retain the semantic content of the original images after style transfer. It is implemented using a VGG19 network pretrained on ImageNET.
    \item \textbf{Expert Model}: A task-specific model is trained to perform recognition on the stylized source training sets generated by DRANet. For character classification, a ResNet-34 \textit{cf.} Section \ref{sub_sec_cnn} is employed, whereas for HTR, a CRNN \textit{cf.} Section \ref{sub_sec_crnn} architecture is used.
\end{itemize}
\Florent{Although both the Encoder and the Expert Model adopt a ResNet-34 backbone, they are instantiated as separate networks with independent parameters.}

\section{Experiments}\label{sec_expe}
In this section, we evaluate the proposed pipeline for both character classification and HTR through a series of experiments. We first describe the datasets used for both source and target domains, including both real and synthetic data. We then detail the training procedure and finally compare our results against state-of-the-art methods to demonstrate the effectiveness and generality of the proposed approach.

\subsection{Character classification}
\subsubsection{Datasets}
\label{sec_data_character_classif}
\begin{figure}[t]
\centering  
    \includegraphics[width=\linewidth]{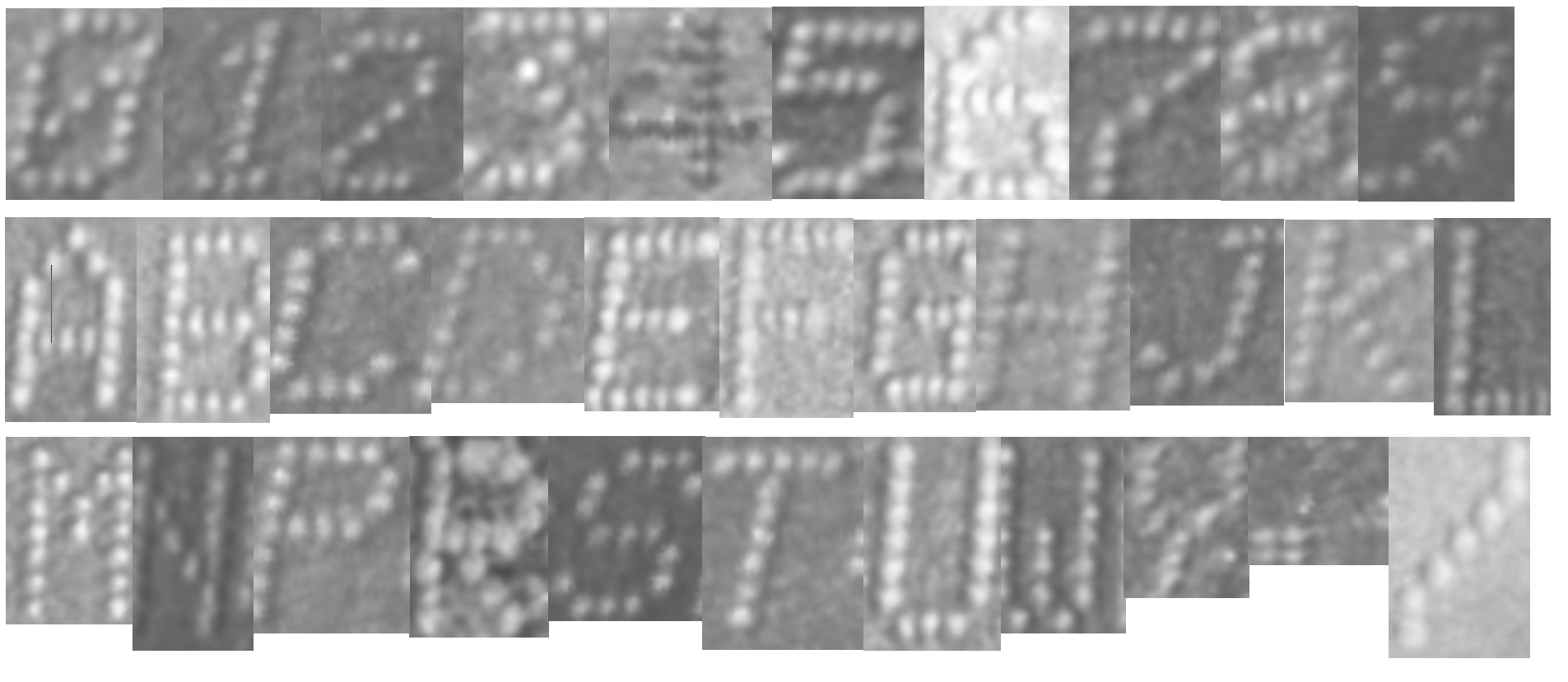}
\caption{Examples of data in Safran-MNIST-DLS for each class.}
\label{fig_data_dagecc}
\end{figure}

To evaluate the impact of DA on a character classification task, we use the Safran-MNIST-DLS dataset from Track 2 of DAGECC~\cite{DAGECC}.
This dataset contains 32 classes depicting digits, alphabetic characters, and symbols. Figure~\ref{fig_data_dagecc} illustrates examples of this dataset for each class.

\Florent{For labeled source data, we use EMNIST~\cite{EMNIST} (digits and capital letters) and synthetic data described in Section~\ref{sec_synthesis}. Table~\ref{tab_db_dagecc} describes the number of characters for each split of each dataset. The EMNIST validation comes from the original train split.} 

\begin{table}[b]
\caption{Number of characters per split for each dataset used in this work for the classification task.}
\begin{tabular*}{\linewidth}{l|r|r|r}
\hline
\textbf{Dataset}            &  \textbf{Train}   & \textbf{Validation}      & \textbf{Test} \\ \hline
Safran-MNIST-DLS            & 9 314             & 862               & 3 448         \\ 
EMNIST$^*$                  & 217 076           &  54 270           & 71 346        \\
Synthetic                   & 190 000           &  47 500           &  47 500          \\ \hline
\multicolumn{4}{l}{\small $^*$digits and upper cases only.} \\
\end{tabular*}
\label{tab_db_dagecc}
\end{table}

\subsubsection{Synthesis for Character Classification}\label{sec_synthesis} 

Our pipeline can integrate synthetic data \Florent{as part of the labeled source data} into step 1 as illustrated in Figure~\ref{fig_overview_method}.
This is especially important when the unlabeled target data differs from the labeled source data. EMNIST is the only available dataset that provides a large amount of labeled character data, to the best of our knowledge, although it constitutes a relatively basic dataset.

\begin{figure}[t]
\centering  
    \includegraphics[width=\linewidth]{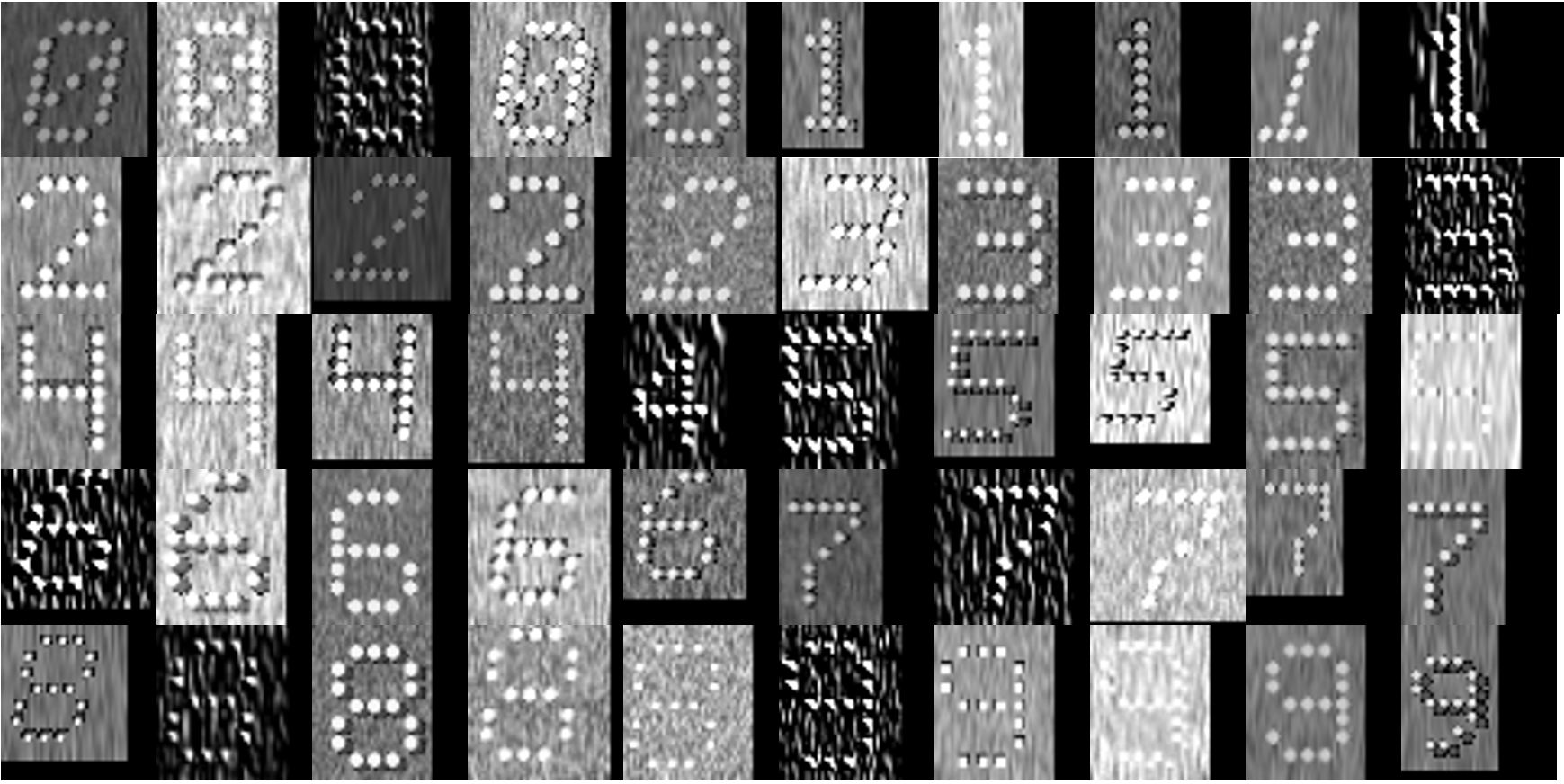}
\caption{Examples of synthetic character images for digit classes.}
\label{fig_ex_synthesis}
\end{figure}

In this section, we describe our synthesis to create data in a character classification context that looks like the Safran-MNIST-DLS dataset described in Section~\ref{sec_data_character_classif}.
The target dataset is composed of 32~classes representing digits, Latin uppercase letters (minus letters that look like a digit), and the dot and slash symbols. Several examples of generated digit classes are illustrated in Figure~\ref{fig_ex_synthesis}.
We select three fonts with a dot style close to the target data. For each font and for each class, we generate N images. We set empirically $N = 2000$. 
The following method is used to generate each character:
\begin{itemize}
    \item Choose a random size.
    \item Draw character in white color.
    \item Draw character in black color with a random offset in order to simulate a shadow.
    \item Randomly apply a shear transformation.
    \item Add a background with Gaussian blur and a random gray value.
    \item Randomly apply embossed effect.
    \item Randomly apply brightness augmentation.
    \item Downscale the image if its size exceeds 48 pixels.
    \item Pad with black color to reach 48 × 48 pixels.
\end{itemize}

All parameters of the synthesis were found empirically.
The corresponding implementation code is available in the Git repository accompanying this article. 
The impact of synthetic data is shown in the results of Section~\ref{sec_expe_quanti_char}.

\subsubsection{Preprocessing and Data Augmentation}
All images are resized to 48 × 48 pixels, which approximately matches the average width and height of the Safran-MNIST-DLS samples. Images larger than this size are downscaled, while smaller ones are padded with black pixels to maintain consistency. 
We use brightness, contrast, sign flipping, Gaussian blur, erosion, dilation, Gaussian noise, and Möbius~\cite{Mobius} data augmentation.

\subsubsection{Training and Evaluation}
We use the ResNet-34 defined in~\cite{ResNet} initialized with ImageNet-pretrained weights as the expert model for the classification task. 
In the supervised step, the expert model is trained with a batch size of 1024 with Adam optimizer with a learning rate of 1e-4 during 150 epochs, except for the case where the model is only trained with Safran-MNIST-DLS data; in this case, the model is trained during 2000 epochs.
The learning rate is divided by 10 at epochs 100 and 1000.
The center loss is always computed from epoch 80.
The best model is the one with the best Macro F1-score on the validation data. This model is used to evaluate the test set. Note that we include Safran-MNIST-DLS validation data only for the configuration where we used the label in the training.

We evaluate performance using the Macro F1-score on the test set, following the protocol defined in the DAGECC competition~\cite{DAGECC}.
The Macro F1-score takes into account the imbalanced nature of the target dataset.
We trained 3 models and computed the mean and the standard deviation.

For steps 2 and 3 of our pipeline in Figure~\ref{fig_overview_method}, the source data is EMNIST~\cite{EMNIST} (digits and capital letters) combined with the synthetic data, and the target data is the unlabeled Safran-MNIST-DLS train dataset.
All experiments were conducted on an NVIDIA A100 GPU with a batch size of 32. The DRANet framework was optimized using the Adam optimizer, with learning rates of 1e-4 for the generative components, 4e-4 for the task-specific classifiers, and 6e-4 for the segmentation modules. The overall objective function comprised multiple weighted loss terms: the adversarial loss with a coefficient of 0.12, the content preservation loss with a weight of 1.0, the style loss with a weight of 0.5, the consistency loss with a weight of 2.0, and the reconstruction loss with a weight of 0.2. 
\Florent{These hyperparameters were selected following the same protocol: for each experimental setting, we start from the original configuration defined in \cite{DRANet}, inspect the training loss curves to assess the relative scale and convergence behavior of each loss term, and adjust their weights to balance optimization dynamics until stable, well-balanced training is achieved.}

\subsubsection{Quantitative Results}
\label{sec_expe_quanti_char}

We first perform an ablation study to analyze the impact of different learning strategies, followed by a comparison of our method with state-of-the-art approaches.

Table \ref{tab_ablation_dagecc} shows the result of our ablation study. The first line presents the results when we train the expert model in a supervised manner with the target data. 
This case from Safran-MNIST-DLS can be seen as an upper bound. When we only use the EMNIST data, there is not enough data and it performs poorly (row 2). The quality of the data available by synthetic production improves the results (rows 3 \& 4). So, we start from this setting (EMNIST + synthetic as source data) to apply the DA, and we see another considerable improvement of 24 \Florent{percentage points} in the Macro F1-score (last row).
The model benefits from pretraining on additional datasets, especially synthetic data, when the target labels are not available, highlighting the importance of designing a high-quality synthetic dataset. While the DA model performs slightly below the upper bound, it reaches comparable accuracy without any labeled target data, demonstrating the practical value of our approach. The standard deviation across 3 runs is around 1, indicating stable and reliable performance.
Overall, these results confirm that our pipeline effectively leverages unlabeled data to significantly improve model performance while reducing the reliance on costly annotations.
Regarding the inference speed, our method can process around 300 images per second with a CPU (13th Gen Intel(R) Core(TM) i7-1355U) and around 13 000 images per second on a GPU (NVIDIA A100).

\begin{table*}[t]
\centering
\caption{Ablation study on supervised configuration and impact of DA on the Safran-MNIST-DLS test set with Macro F1-score ($\uparrow$).}
\begin{tabular}{lcc}  
\hline
\textbf{Training data supervised}      & \textbf{DA}   & \textbf{Test Macro F1-score (\%) $\uparrow$} \\ \hline
Safran-MNIST-DLS                       &               & 84.98 $\pm$ 1.11      \\ 
\hline
EMNIST                                 &               & 8.95 $\pm$ 1.50        \\
Synthetic                              &               & 48.53 $\pm$ 2.67       \\
EMNIST, Synthetic                      &  ~~~~~~~~~~~~~~~~~~ & 51.71 $\pm$ 3.50        \\
EMNIST, Synthetic                      &  $\checkmark$ & \textbf{76.17} $\pm$ 1.07       \\
\hline
\end{tabular}

\label{tab_ablation_dagecc}
\end{table*}


Table \ref{tab_soa_dagecc} compares our method against the state-of-the-art from the DAGECC competition~\cite{DAGECC}. Note that the top 3 best methods of the state-of-the-art do not use unlabeled data.
Our method outperforms the state-of-the-art by \Florent{11 percentage points} in Macro F1-score on the test set, highlighting the interest in using unlabeled target data to improve model performance.
Additionally, we can note our method does not add extra inference time.

\begin{table*}[t]
\centering
\caption{Comparison of our method with the state-of-the-art results on Track 2 of the DAGECC competition on the Safran-MNIST-DLS test, with Macro F1-score ($\uparrow$).}
\begin{tabular}{lrrc}  
\hline
\textbf{Method}                     & & & \textbf{Test Macro F1-score (\%) $\uparrow$} \\ \hline
DAGECC - JasonMendoza2008           & & &  14       \\ 
DAGECC - Machine Learning Group LTU & & &  44       \\
DAGECC - Raul                       & & &  52          \\ 
DAGECC - Deep Unsupervised Trouble  & & &  60       \\ 
DAGECC - Team Deng                  & & &  65        \\ 
Ours                                & & &  \textbf{76.17} $\pm$ 1.07        \\ 
\hline
\end{tabular}

\label{tab_soa_dagecc}
\end{table*}

 
\subsubsection{Qualitative Results}
\begin{figure*}[t]
\centering  
    \includegraphics[width=0.48\linewidth]{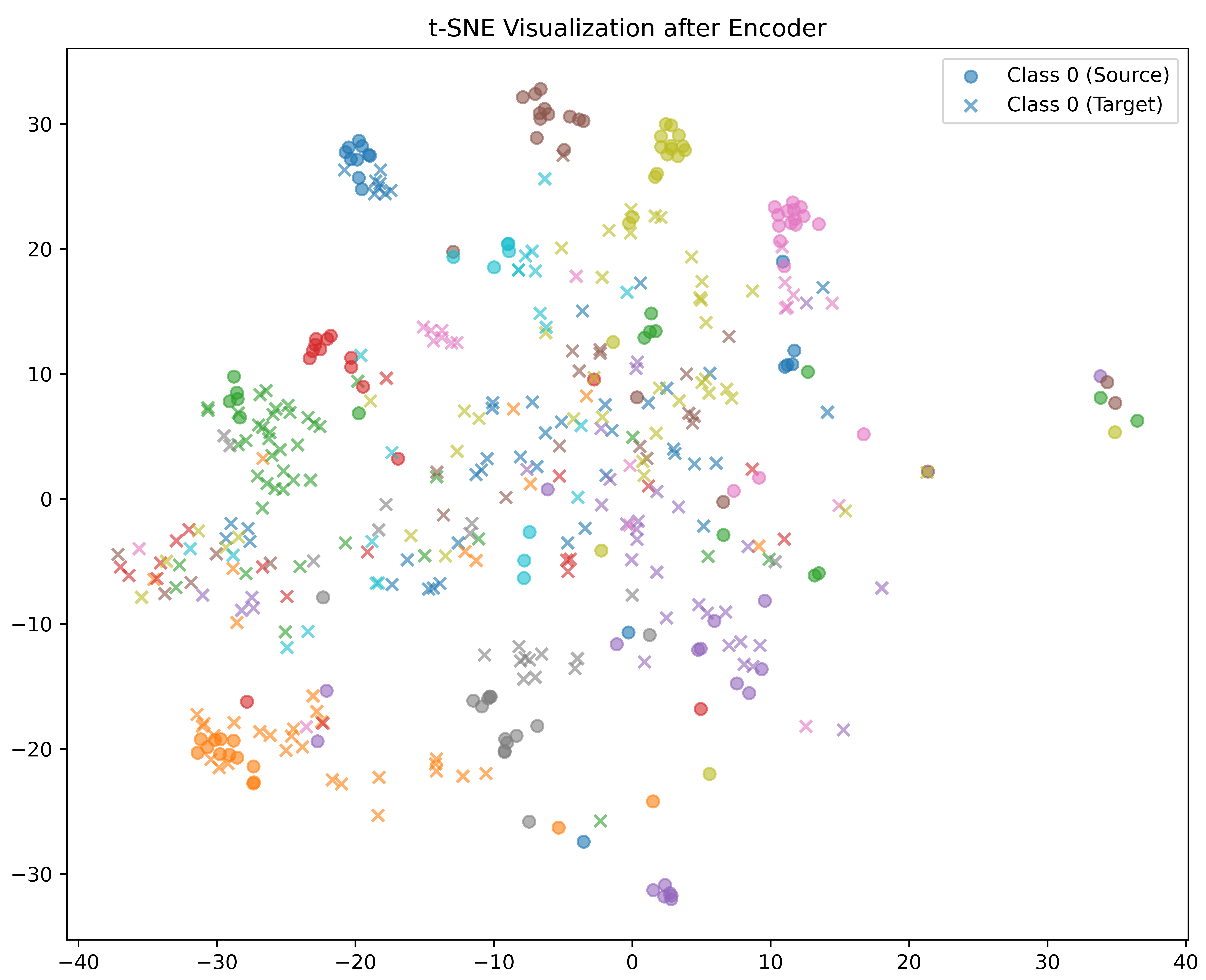}
    \includegraphics[width=0.48\linewidth]{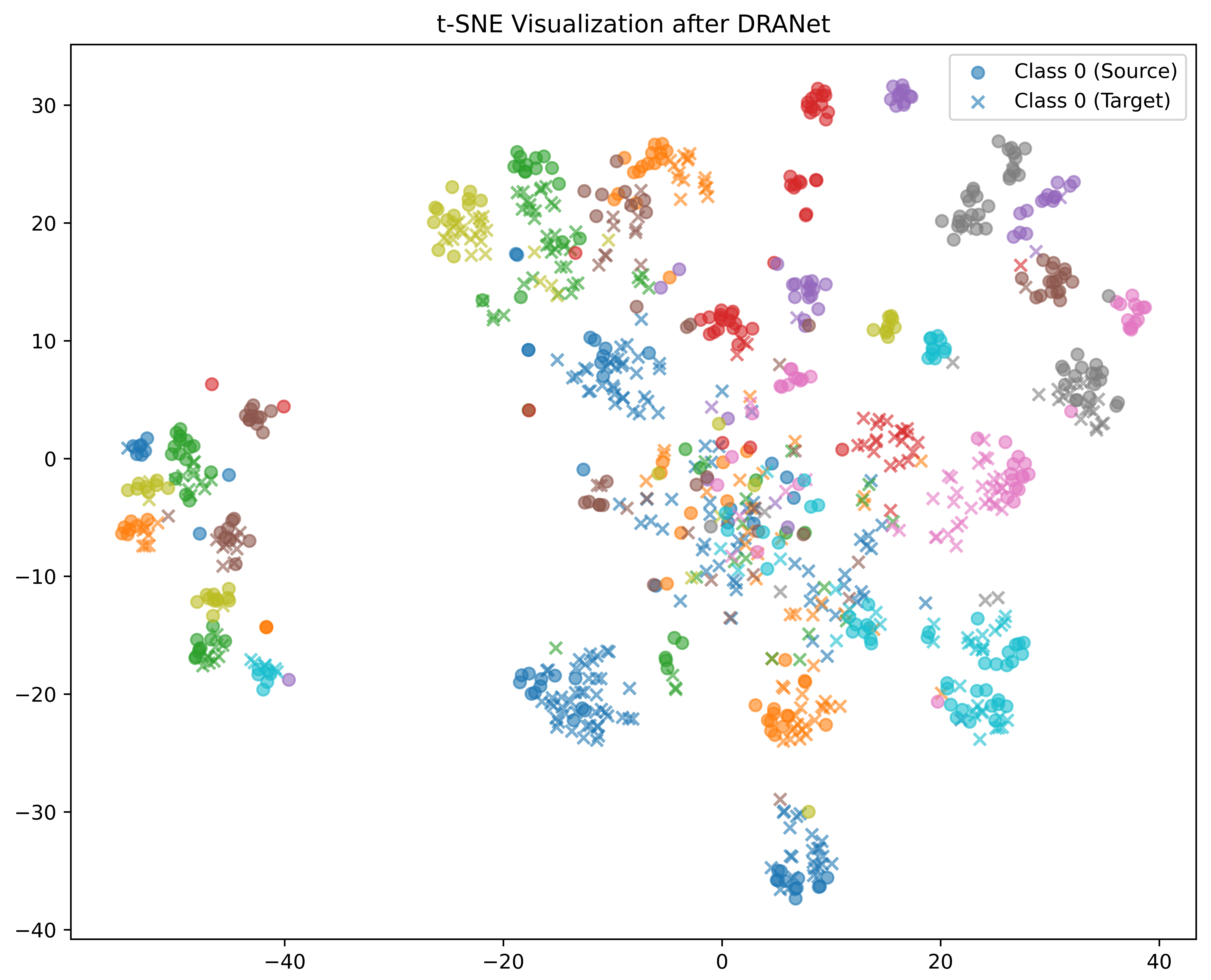}
\caption{Visualization in 2D feature space of the expert model with T-SNE with source and target data. }
\label{fig_TSNE}
\end{figure*}

Figure \ref{fig_TSNE} illustrates the feature space at the encoder output (\textit{cf.} Section~ \ref{sub_sec_crnn}) and after the DA pipeline. This reveals the impact of the DA method. Before the adaptation, the source features exhibit clear clustering by class, whereas the target features are dispersed and don't form clear clusters, highlighting the domain shift. After applying DRANet, both source and target show better class wise clusters, that indicate that the method successfully aligns the target distributions with the source.  

\begin{figure}[t]
\centering  
    \includegraphics[width=\linewidth]{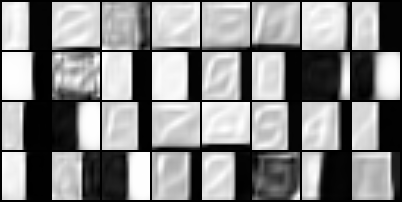}
\caption{Example of a synthetic image generated by the domain adaptation pipeline.}
\label{DA_syn_image}
\end{figure}

\Florent{In addition, Fig.~\ref{DA_syn_image} illustrates examples of images generated by the proposed domain adaptation pipeline using unlabelled target data. While the synthetic images do not perfectly match the visual quality of the original source data, they successfully capture key target domain attributes such as illumination and appearance variation while preserving the underlying source domain semantic content.}

\subsection{Handwritten line text recognition}
Now that we have illustrated how DA can improve classification on isolated characters,
we turn our attention to handwritten text recognition.
In this section, we evaluate the performance of our pipeline for HTR. Unlike isolated character classification, HTR requires recognizing entire handwritten lines and has more than 32 classes, making it a more complex sequence prediction problem. Our goal is to demonstrate that the proposed DA pipeline, initially validated on character classification, can be effectively extended to HTR, thereby confirming its generality and robustness. We first describe the datasets used for both source and target domains for the HTR task. We then detail the training procedure, and finally compare our approach with state-of-the-art methods.

\subsubsection{Datasets}
\begin{figure}[t]
\centering  
    \includegraphics[width=\linewidth]{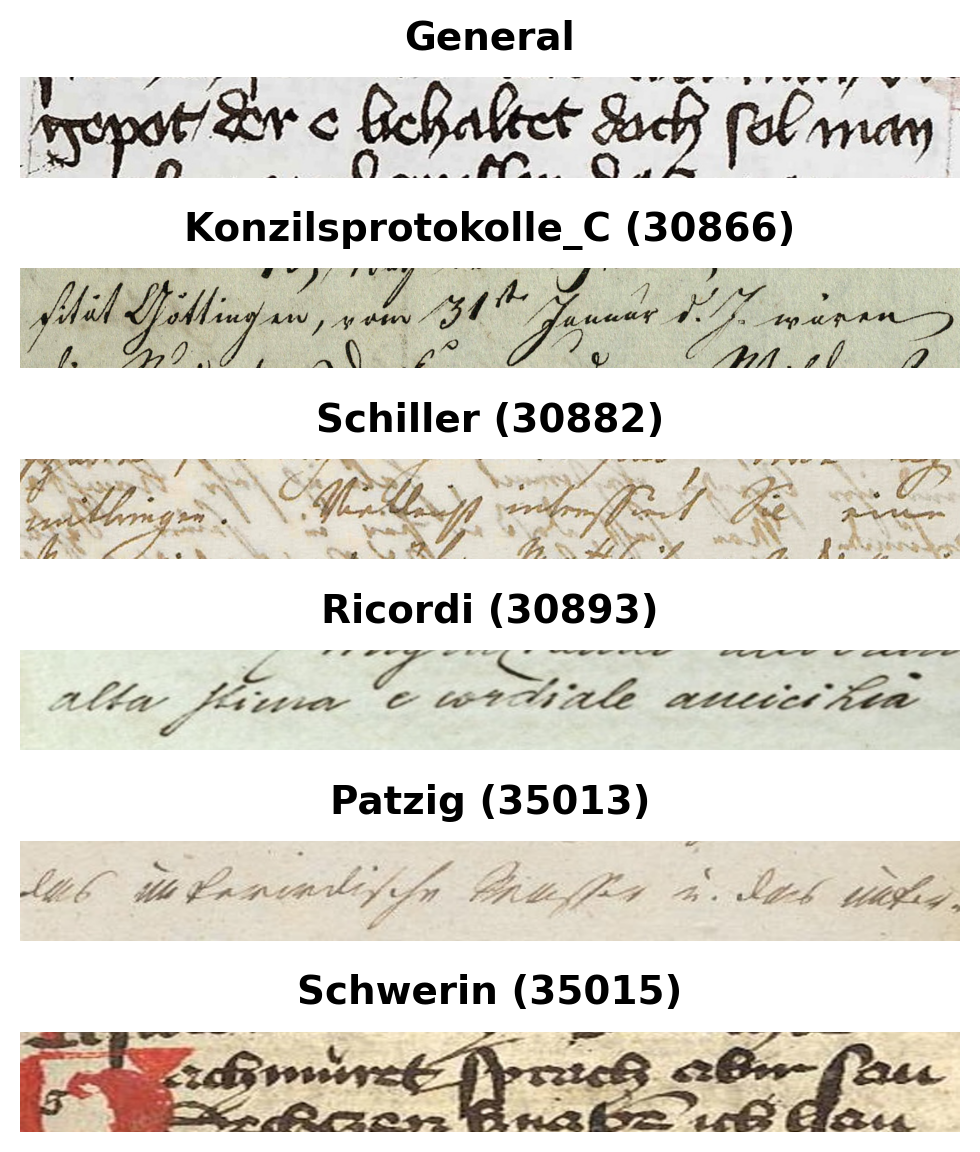}
\caption{Examples of data in READ 2018 for the general and the five specific target datasets.}
\label{fig_data_read}
\end{figure}

The ICFHR 2018 READ historical handwritten dataset~\cite{READ2018} provides a challenging benchmark for evaluating models trained with limited annotated data. It consists of a generic document sample collection which is to be used for training, then samples from five specific documents to be used for adaptation and testing.
The general dataset has  11,903 annotated text-line images from 17 documents (\textit{cf.}~Figure~\ref{fig_data_read}) in German, English, Swedish, and Danish, which can be used to pre-train a model. The main objective is then to adapt this pre-trained model to five additional specific documents while minimizing the use of additional labeled data. The specific dataset contains five documents: Konzilprotokolle C, Schiller, Ricordi, Patzig, and Schwerin. All documents are written in German except Ricordi, which is in Italian, introducing an additional challenge since the general pre-training dataset contains no Italian samples.

The dataset is designed to evaluate adaptation capabilities using varying amount of labeled data per document: 0, 1, 4 or 16 pages. The case with 0 labeled pages is particularly relevant for evaluating unsupervised DA, as it demonstrates the model's ability to generalize to a new domain without requiring any additional annotations beyond the general dataset.
Evaluation is performed on a separate test set with samples from each specific document.

\subsubsection{Preprocessing and Data Augmentation}
We set the fixed size of an image to a height of 160 and a width of 1570 which corresponds approximately to the mean values of the general training data. 
The larger images are downscaled. We also added a left and right padding of 64.
After preprocessing, all the images have the same size.
We used affine, grid distortion, elastic transform, erosion, dilation, brightness gamma as well as random noise data augmentation as detailed in~\cite{Retsinas22}.

\subsubsection{Training and Evaluation}
The CRNN model is trained on an NVIDIA A100 GPU for 1000 epochs using the Adam optimizer with a learning rate of 5e-4, and a batch size of 32. From epoch 700 the learning rate is decreased by a factor of ten. The center loss is computed from epoch 500 with a weight of 0.55.

Recognition performance is evaluated with the usual Character Error Rate (CER) and Word Error Rate (WER). In each experiment, we train the expert model three times and calculate both the mean and standard deviation. The expert model with the smaller CER on the validation set is selected as the best expert model and is evaluated on the test set. 

In steps 2 and 3 of our pipeline, shown in Figure~\ref{fig_overview_method} the source domain corresponds to the READ 2018 general dataset, and the target data is respectively each of the five READ 2018 specific datasets (16 pages without labels).
All experiments were conducted on an NVIDIA A100 80 GB GPU with a batch size of~8. For DA, it is important to train with sufficiently large and representative batches, since aligning the source and target distributions requires a diverse set of samples in each iteration. However, GPU memory constraints often limit the direct use of large batch sizes. To overcome this limitation, we employ the gradient accumulation technique, which simulates larger batch sizes without increasing memory usage. In practice, we accumulate gradients over several smaller mini-batches before updating the model parameters, which enables us to reach an effective batch size of 64. This strategy provides a better statistical representation of both source and target domains during training, leading to more stable domain alignment and improved adaptation performance.  
We used two sets (set A and set B) of hyperparameters to train the DRANet model. The first one is used for the Konzil C and Schiler datasets, and the other one for the three remaining datasets.
The DRANet framework is optimized using the Adam optimizer, with the parameter shown in Table \ref{tab:hyperparameters}.
These hyperparameters were carefully chosen to ensure stable training of the DRANet model while promoting both effective semantic content preservation and stylistic adaptation in the generated images. 
\begin{table}[h]
\centering
\caption{\Florent{Hyperparameters for DRANet framework}}
\begin{tabular}{lcc}
\hline
\textbf{Parameter} & \textbf{Set A} & \textbf{Set B} \\
\hline
\multicolumn{3}{l}{\textit{Learning Rates}} \\ \hline
Generative components & \(4 \times 10^{-5}\) & \(2 \times 10^{-4}\) \\
Task-specific classifier & \(3.5 \times 10^{-4}\) & \(4 \times 10^{-4}\) \\
Segmentation module & \(5 \times 10^{-4}\) & \(6 \times 10^{-4}\) \\
\hline
\multicolumn{3}{l}{\textit{Loss Weights}} \\ \hline
Adversarial loss & 0.14 & 0.12 \\
Content preservation loss & 1.3 & 1.0 \\
Style loss & 0.2 & 0.5 \\
Consistency loss & 2.2 & 2.0 \\
Reconstruction loss & 0.05 & 0.2 \\
\hline
\end{tabular}
\label{tab:hyperparameters}
\end{table}


\begin{table*}[t]
\centering
\caption{Results of the CRNN with supervised training and our DA pipeline for each scenario. Recognition is evaluated on the test set with Character Error Rate (CER) (lower value is better).  \Simon{Our expert columns correspond to the number of labeled pages used for training.}}
\label{tab_ablation_read2018}
\small
\begin{tabular}{lcccc|c}
\hline
\multirow{2}{*}{\textbf{Document}}  & \multicolumn{4}{c|}{Our expert (CRNN)} & CRNN + DA  \\
                                    & 0 pages  & 1 pages & 4 pages & 16 pages     &  using 16 unlabeled pages                            \\
\hline
Konzil C   & $9.89  \pm 0.10$ & $7.18 \pm 0.33$ & $4.73 \pm 0.15$ & $3.69 \pm 0.06$ & $\textbf{3.10} \pm 0.76$ \\
Schiller   & $16.75 \pm 0.35$ & $12.70 \pm 0.19$ & $10.57 \pm 1.03$ & $6.94 \pm 0.17$ & $\textbf{1.89} \pm 0.11$ \\
Ricordi    & $25.35 \pm 0.51$ & $22.64 \pm 0.59$ & $15.16 \pm 0.40$ & \textbf{9.45} $\pm 0.45$ & $14.08 \pm 9.60$ \\
Patzig     & $35.63 \pm 0.85$ & $23.06 \pm 0.80$ & $16.06 \pm 0.30$ & 9.76 $\pm 0.16$ & \textbf{8.54} $\pm 3.56$ \\
Schwerin   & $28.74 \pm 1.37$ & $10.25 \pm 0.20$ & $6.04 \pm 0.06$ & 3.61 $\pm 0.12$ & \textbf{2.70} $\pm 2.25$ \\
\hline
\end{tabular}
\end{table*}

\subsubsection{Quantitative Results}
In this section, we first conduct an ablation study to evaluate the effects of different learning strategies in both the expert model and the DA pipeline, and then compare our approach with state-of-the-art methods.

Table \ref{tab_ablation_read2018} shows the result of our ablation study. 
The first part (left) of Table \ref{tab_ablation_read2018} reports the result of fully supervised learning for the CRNN (our expert model) trained only on the general READ 2018 dataset (0 pages), then finetuned on (1, 4 or 16 pages) labeled from each specific dataset.
The second part (right) represents the results of our complete pipeline, which combines a CRNN pretrained on the generic dataset (0 pages) with the DA method applied using each specific dataset (unlabeled 16 pages) of READ 2018.

We observe a considerable improvement in the CER when DA is applied using no labeled pages compared to the 0 labeled page setup, which demonstrates the value of DA in adapting a model to specific sets without the need for additional annotation. However, DA performs least effectively on the Ricordi set, which is written in Italian, unlike the predominantly German datasets. This suggests that the benefits of DA are reduced when the data distribution (text, language and writing style) differs significantly from that of the source domain.
We can also note a considerable standard deviation for this dataset as well as for Patzig and Schwerin.
When comparing our approach to our expert model trained on fully supervised setting with 16 labeled pages, our pipeline achieves better performance using no labeled pages than the expert with 16 annotated pages on four of the five datasets, confirming the robustness and consistency of our DA strategy across diverse conditions.

We next compare our pipeline against the state-of-the-art method \cite{barrere24} in Table \ref{tab_sota_read2018}.
Even if our expert model is slightly less accurate compared to the state-of-the-art with 16 labeled pages, we observe the same behavior.
Our pipeline outperforms the state-of-the-art when no specific data are used for the training (0 pages) across the five datasets and when 16 pages are used for the Schiller dataset.

\begin{table*}[t]
\centering
\caption{Results of our pipeline against the state-of-the-art for supervised training on each scenario. Recognition is evaluated on the test set showing Character Error Rate (CER) (lower value is better). \Simon{The SOTA columns correspond to the number of labeled pages used for training.}}
\label{tab_sota_read2018}
\small
\begin{tabular}{lrrrr|r}
\hline
\multirow{2}{*}{\textbf{Document}} & \multicolumn{4}{c|}{SOTA \cite{barrere24}} & Our pipeline \\
                          & 0 pages  & 1 pages & 4 pages & 16 pages &    CRNN + DA \\
\hline
Konzil C  & $10.54$ & $6.81 \pm 0.73 $ & $4.33 \pm 0.32$ & \textbf{2.95} $\pm 0.12$ & $3.10 \pm 0.76$ \\
Schiller   & $21.20$ & $14.92 \pm 0.12$ & $11.86 \pm 0.26$ & $7.64 \pm 0.31$ & \textbf{1.89} $\pm 0.11$ \\
Ricordi    & $22.96$ & $16.72 \pm 0.89$ & $12.77 \pm 0.13$ & $\textbf{8.41} \pm 0.29$ & $14.08 \pm 9.60$ \\
Patzig     & $29.37$ & $17.46 \pm 0.60$ & $12.10 \pm 0.07$ & $\textbf{7.67} \pm 0.23$ & 8.54 $\pm 3.56$ \\
Schwerin   & $30.09$ & $9.37 \pm 0.70$ & $4.75 \pm 0.13$ & \textbf{2.63} $\pm 0.08$ & 2.70 $\pm 2.25$ \\
\hline

\end{tabular}
\end{table*}

These results demonstrate the strength of DA in transferring knowledge from the source to the target domain.
The Ricordi set represents the scenario where our pipeline is least effective, likely due to its different language and consequently distinct word and letter distributions, which limit the transferability of knowledge from the source domain \Florent{in the linguistically sensitive LSTM}.
Overall, these results confirm that our pipeline effectively leverages unlabeled data to significantly improve model performance while reducing the reliance on costly annotations.
Regarding the inference speed, our method can process around 3 line images per second with a CPU (13th Gen Intel(R) Core(TM) i7-1355U) and around 1~700 line images per second on a GPU (NVIDIA A100).

\section{Conclusion}
\label{sec_conclu}

We propose a novel domain adaptation–based pipeline and evaluate it on both character classification and historical handwritten line recognition tasks, 
\Simon{illustrating the generality of our method.}

Our pipeline operates in three main steps. In the first step, we pretrain an expert model on labeled source data, optionally augmented with synthetically generated samples.
In the second step, we apply unsupervised DA to leverage unlabeled target data, refining the model to the specific characteristics of the target domain. In the third step, the expert model is trained with source data in the target style.
We can also note that our method does not add computation time for the inference; the method only adds computation time during the training.

In our evaluation, we compare the performance of our pipeline using DA against both fully supervised settings and existing state-of-the-art methods. Our pipeline outperforms existing state-of-the-art approaches, establishing new benchmarks on the character level (Safran-MNIST-DLS) and the line level (READ 2018) datasets. By demonstrating the effectiveness of our pipeline on the character level and extending it to the line level, we hence highlighted its potential for extension to more complex scenarios, such as page-level recognition. 

\backmatter





\bmhead{Acknowledgements}
This work was supported by the Kempestiftelserna grant JCSMK23-0109.
The computations were enabled by the Berzelius resource provided by the Knut and Alice Wallenberg Foundation at the National Supercomputer Centre.



\bibliographystyle{ieeetr-etal}  
\bibliography{biblio}

\end{document}